# A prototype system for handwritten sub-word recognition: Toward Arabic-manuscript transliteration

Reza
FARRAHI MOGHADDAM
Synchromedia Laboratory
École de Technologie Supérieure
Montreal, (QC), H3C 1K3 Canada
reza.farrahi@synchromedia.ca
imriss@yahoo.com

Mohamed CHERIET
Synchromedia Laboratory
École de Technologie Supérieure
Montreal, (QC), H3C 1K3 Canada
mohamed.cheriet@etsmtl.ca

Thomas MILO
DecoType
Amsterdam, Netherlands 1072 LL
tmilo@decotype.com

Robert WISNOVSKY
Institute of Islamic Studies
McGill University
Montreal (QC), H3A 1Y1 Canada
robert.wisnovsky@mcgill.ca

## ABSTRACT
A prototype system for the transliteration of diacritics-less Arabic manuscripts at the sub-word or part of Arabic word (PAW) level is developed. The system is able to read sub-words of the input manuscript using a set of skeleton-based features. A variation of the system is also developed which reads archigraphemic Arabic manuscripts, which are dot-less, into archigraphemes transliteration. In order to reduce the complexity of the original highly multiclass problem of sub-word recognition, it is redefined into a set of binary descriptor classifiers. The outputs of trained binary classifiers are combined to generate the sequence of sub-word letters. SVMs are used to learn the binary classifiers. Two specific Arabic databases have been developed to train and test the system. One of them is a database of the Naskh style. The initial results are promising. The systems could be trained on other scripts found in Arabic manuscripts.

## Categories and Subject Descriptors
I.7.5 [**Document and Text Processing**]: Document Capture—*Document analysis, Graphics recognition and interpretation*; I.2.6 [**Artificial intelligence**]: Learning

## 1. INTRODUCTION
The special feature of Arabic manuscripts is the cursive nature of their scripts [1], which means that they are more oriented toward sub-words (letter-blocks or connected-components: "CCs") than words. This is particularly true of pre-modern manuscripts, in which there is no measurable difference in the distances between sub-words and words. It is worth noting that sub-words (or part of Arabic word: PAW) in the Arabic language are any set of letters (letter-blocks) which are disconnected at the pixel level. To add to this complexity, the shapes of letters change according to their position within a sub-word (that is, each letter has various allographs). The presence of cavities inside the shapes further increases the complexity of these scripts, and special features are required to describe them. On top of all this, the high degree of intra-script variation makes the task of achieving a single solution for all Arabic scripts extremely difficult. The main challenge confronting these methods is the increase in their complexity when moving from a low-complexity database, such as city names, to full set of words in the language [2, 3].

In this work, we use sub-words in order to skip the line- and word-segmentation problem encountered in pre-modern Arabic manuscripts. After recognizing All sub-words of a manuscript, its words should be reconstructed which is beyond the scope of this work. We provide a complete recognition chain at the sub-word level. It works directly with the sub-words, and does not try to break them into character segments. Therefore, we call it an Optical Shape Recognition (OSR) system. The system uses a novel concept we call it the binary descriptor paradigm (we have also used the binary problem paradigm notation for it [4]). In this paradigm, a set of overlapping binary descriptors allows us to classify all sub-word classes without segmenting the sub-words into characters. To achieve this, we consider membership functions that arise from creating a new representation of sub-words in terms of their letters. For example, a sub-word "bkt", which is usually is represented by an ordered vector ("b", "k", "t"), will be represented by a set of binary descriptors $\{P_\xi\}_\xi$ where $\xi$ counts for all letters, and $P_\xi$ is one of the binary descriptors used to describe the sub-words. As can be seen from the definition, the new representation is order-free, and therefore each descriptor $P_\xi$ can be processed independently. It is worth noting that, although there is no order in the set of binary descriptors, they can carry order information within themselves. For example, a binary descriptor could be *if the second letter of the sub-word is letter "m" or not*. For the sub-word "bkt", this binary descriptor will give 0, because the second letter is "k". More discussions of the binary descriptors paradigm is provided in section 3.1. It is worth noting that this concept has been previously proposed in [4]. However, in that work, only the general idea was discussed, and as a proof of concept, a small number of binary descriptors (problems) which had more than 1000 positive samples was considered and learned with a promising error range. Therefore, no attempt was made to recover the sequence of letters of a sub-word. In this work, not only is a complete set of binary descriptors considered even if the number of positive samples is very low, the system provides a set of candidate sequences of letters for each sub-word by combining the values of the binary descriptors. In order to obtain the full text, sub-words should be combined using language-level analysis and word distributions. This step is beyond this work, and will be addressed in future research.

A schematic flow diagram of the proposed OSR system is shown in Figure 1. The input document images are pre-processed, and then the sub-words are extracted easily by identifying CCs. The feature vector of the sub-words is calculated according to their skeleton and some *a priori* information, such as the average stroke width (see section 2.1). The binary labels of each sub-word are obtained using trained machines, and then are combined to generate candidate sequence of each sub-word letters. Using a dictionary of sub-words, the set of candidate sequences is pruned, and the final set of candidate sequences for each sub-word is provided as the output of the system. We use the terminology "string" for the sequence of sub-word letters to avoid confusion with the labels of the binary descriptors. Also, it is worth noting that by "Arabic-scripted language", we refer to all languages whose scripts are based on the Arabic script, including not only Arabic but the Persian, Urdu and Ottoman-Turkish languages. In other words, we do not limit the scope of letters to a specific language. However, the scope will be automatically limited in the training stage for each language and selected script and style. Script, style and language identification steps are ignored in this work. By training our system on different scripts, styles and languages, and adding associated identification steps, it can be used to read manuscripts from those languages and scripts. Databases are the building blocks of recognition systems [3–7]. For example, a database for the recognition of legal amounts and Arabic sub-words on Arabic checks has been developed [6] which contains 1547 legal amounts and 23325 sub-words. We used two databases in this work: i) an Arabic language database created from a real historical manuscript, and ii) a synthesized archigraphemic-Arabic language database in Naskh style. Archigraphemic-Arabic ignores notation of dots. Therefore, letters with the same archigrapheme, such as bā' ب and tā' ت, appear exactly the same, and are represented by the dot-less bā' ٮ. Because differentiation between these letters needs language-level analysis which is not included in the current system, the output of system will also be in archigraphemes. It is also worth noting that an archigraphemic-Arabic system can be used to recognize normal Arabic manuscripts by stripping out the dots before feeding the manuscript to the system, and then using a dot analysis which recovers Arabic letters from the Arabic archigraphemes using the dot information. A snapshot of the user interface of our system is shown in Figure 2. The user can easily click on a sub-word and the ground-truth sequences and the first-rank recognized sequence will be shown. The databases are discussed further in section 2.

We use support vector machines (SVMs) as the learning machines. They are trained as follows. Having the ground truth sequence of all the sub-words, the labels of the binary descriptors are generated and fed into the SVMs. An optimization of the SVM parameters is also performed. We used two databases that are available to us for training and testing the proposed system. The first is the IBN SINA database [4] and the second is a new database that we have developed using a font system for Arabic scripts[1] [8]. The same procedure was applied to both databases, the main difference being that archigrapheme encoding is used in the second database in place of the grapheme encoding used in the IBN SINA database. An archigrapheme is the bundle of shared features between two or more graphemes, minus their distinctive features (diacritics) [9]. Particularly for the Arabic script, an archigrapheme is a diacritics-less ductus of its associated graphemes [10]. The archigraphemes are shown in Figure 5.

The organization of the paper is as follows. In section 2, more detail on the materials used in the development of the databases is provided. The procedure that we followed for training the SVMs and building the database dictionary is described in section 3. The performance of the whole system on the databases is presented in section 4. Finally, a discussion, our conclusions, and future prospects are provided in sections 5 and 6.

## 2. TWO ARABIC SUB-WORD DATABASES

Two databases were used in this work. The first is the IBN SINA database built based on manuscript images provided by the Institute of Islamic Studies (IIS), McGill University, Montreal. The author of the manuscript is Sayf al-Din Abu al-Hasan Ali ibn Abi Ali ibn Muhammad al-Amidi (d. 1243A.D.). The title of the manuscript is *Kitab Kashf al-tamwihat fi sharh al-Tanbihat* (Commentary on Ibn Sina's [i.e., Avicenna, d. 1037A.D.] *al-Isharat wa-al-tanbihat*). Of all of his philosophical works, Ibn Sina's *al-Isharat wa-al-tanbihat* received the most attention from later philosophers and theologians. The database consists of 51 folios, and contains 20722 sub-words.

The second dataset is based on Arabic Calligraphic Engine (ACE), which is a font-layout engine. ACE is developed to approach Arabic computer typography in complete analogy with pre-typographic text manufacture, and is the proof-of-concept for modern smart-font technology. Currently, only sub-words containing up to 3 letters have been added to the database, which from now on will be called the Naskh-3 database. In contrast to the IBN SINA database, in the Naskh-3 database, archigraphemes are the smallest unit to be recognized. As discussed in the introduction, our aim with this choice was to try an alternative recognition, in which archigraphemes are recognized first, and then they are relabeled with their grapheme in a second round. In this way, the high complexity of scripts can be addressed in two steps. Note that we target only the first step in this work, i.e., the recognition of archigraphemes.

---

[1]Tasmeem: http://www.decotype.com/

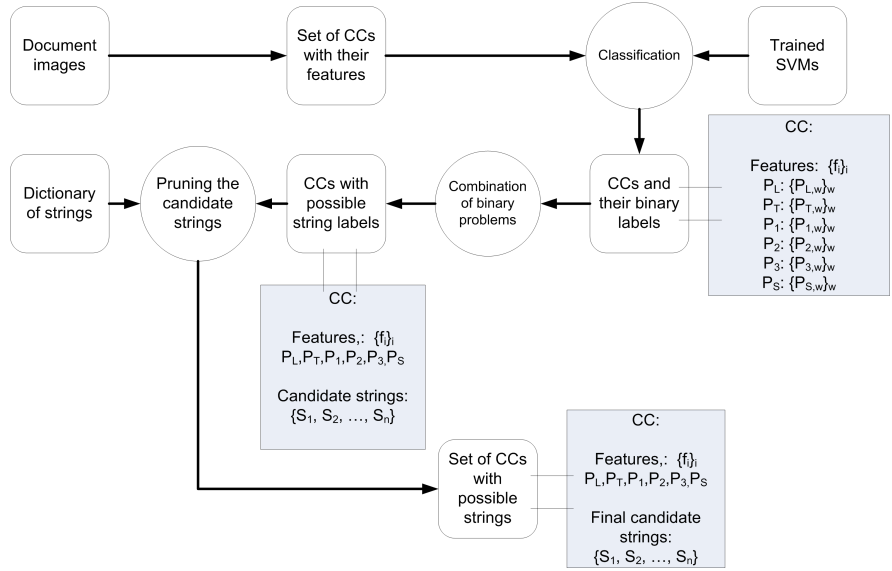

Figure 1: The flow diagram of the proposed OSR system. The structure of data is shown at each step in shaded boxes.

It is worth noting that our databases are diacritics-less, i.e., diacritical marks do not exist in them. In the case of the IBN SINA database, this is due to the nature of the manuscript used. We did not include diacritics in Naskh-3 database, in order to keep it simple. Developing a system that processes manuscripts with diacritics is beyond this work. Also, touching of sub-words is not applicable to our databases because of the high-quality writing hands of their associated document images.

## 2.1 Skeleton-based features

We generate the skeleton-based features of each sub-word. A sample sub-word and its skeleton are shown in Figure 3. Starting with the skeleton of a sub-word, its end points (EPs), branch points (BPs) and dots or singular points (SPs) are identified. Then, some features are assigned to each of these points depending on their connectivity to the others, in order to include the topology of the skeleton in the features. Some other features are assigned as well, to capture the geometrical relationship between the points. These features can be represented as a set of transformations from the skeleton image space to some associated feature spaces. Let us consider $u$ and $u_{skel}$ to be the images of a sub-word (CC) and its skeleton respectively, where $u_{skel} : \Omega_{skel} \to \{0,1\}$, and $\Omega_{skel} = \Omega_u \subset \Omega \subset \mathbb{R}^2$. $\Omega$ is the domain of the whole page that is hosting the sub-word under study. Let us call $T$ the set of our transformations that maps $u_{skel}$ to the proper spaces: $T = \{T_i | T_i : \Omega_{skel} \to (\mathbb{R}^{m_i})^{n_i}, i = 1, \cdots, n_T\}$, where $n_T$ is the number of transforms and $m_i$ depends only on the transformation $T_i$, while $n_i$ depends on the complexity of $u_{skel}$ as well:

$$u_{skel} \xrightarrow{T} v = \{f_i\}_{i=1}^{n_T}, \quad (1)$$
$$(\mathbb{R}^{m_i})^{n_{i,u_{skel}}} \ni f_i = T_i(u_{skel}) = \{\phi_{i,j}\}$$

It is worth noting that the transformation $T$ varies according to the complexity of the sub-word. In other words, the dimensions of the target feature spaces are not constant. The list of the features can be found in Figure 1. The details are as follows:

1. $T_1$ extracts features from BPs:
    - BHoleCon is 1 if the BP is connected to a hole,
    - BEPCon is 1 if it is connected to an EP, and
    - BBPCon is one if it is connected to another BP.

2. $T_2$ extracts features from EPs:
    - EBPCon is 1 if the EP is connected to a BP,
    - EEPCon is 1 if it is connected to another EP, and
    - ERelVertCMEP is positive if it is above the vertical center of mass of the sub-word.

3. $T_3$ extracts dot-related features of a BP:
    - BDotUpFlag is one if there is a dot above the BP, and
    - BDotDownFlag is one if there is a dot below it.

4. $T_4$ extracts dot-related features of a EP: EDotFlag is 1 if there is a dot assigned to the EP.

5. $T_5$ extracts dot-related features of a dot: DRelVertCM-Dot is positive if the dot is above the vertical center of mass of the sub-word.

6. $T_6$ extracts dot-related features of an EP branch:

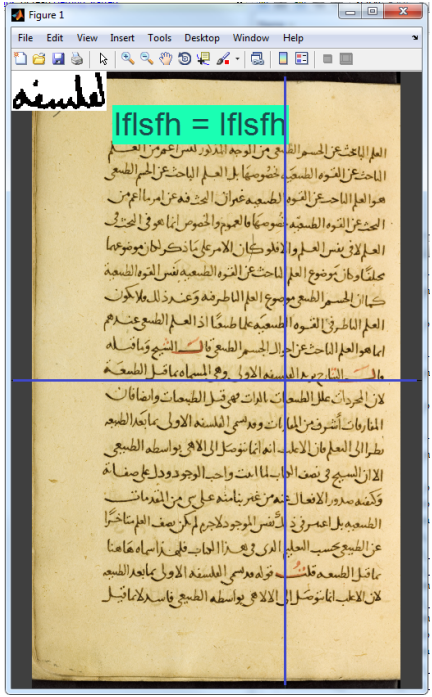

Figure 2: A snap-shot of the system's user-interface. By clicking on the sub-words, their sequences appear on the image in Finglish. Unicode fonts will be integrated to the interface soon.

- ESShapeFlag is 1 if the branch is S-shape,
- EClockwise is positive if it is clockwise,
- EAboveItsBP is 1 if its EP is above its BP, and
- EBelowItsBP is 1 if its EP is below its BP.

Also, 8 global features are assigned which we consider them as $T_G$ (see Table 2). The details are as follows:

1. AR is Aspect ratio.
2. HorizFreq is the number of peaks in the horizontal profile of the sub-word.
3. VertCMRatio is the ratio of the vertical center of mass to the sub-word height.

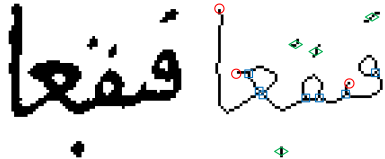

Figure 3: A sample sub-word from the IBN SINA database and its skeleton image. The branch points (blue), end points (red), and singular points (green) are also shown on the skeleton image.

4. # SPs is the number of singular points in the sub-word.
5. Heightratio is the ratio of the sub-word height to the average text height.
6. HoleFlag is 1 if there is a hole in the sub-word.
7. # EPs is the number of end points in the sub-word.
8. DottedFlag is 1 if there is a dot in the sub-word.

In order to have a coherent set of features for all the shapes, a limit on the number of different points, $l_{point}$, is assumed. In this way, if, for example, there are more than $l_{point}$ EPs for a sub-word, then all EPs after $l_{point}$ are dropped. If the number of points is less than $l_{point}$, the rest of the vector will be filled with zeros. As Arabic manuscripts are written from right to left, the first $l_{point}$ points from the right side of a sub-word is considered. In this work, we assume that $l_{point} = 6$, which means that 84 skeleton-based features are assigned to each shape. Adding the 8 global features, this brings the total number of features assigned to each shape to 92: $x_i = \{x_{i,\omega}\}_{\omega=1}^{92}$, where $x_i$ is one of the features vectors, and $\omega$ is the index. A typical feature vector is shown in Table 3. It starts with the global features that are followed by the six occurrences of each $T_i$, $i = 1, \cdots, 6$.

| $T_G$ | | | |
|---|---|---|---|
| 1 | AR | 5 | Heightratio |
| 2 | HorizFreq | 6 | HoleFlag |
| 3 | VertCMRatio | 7 | # EPs |
| 4 | # SPs | 8 | DottedFlag |

Table 2: The 8 global features of a sub-word.

Here, we provide a short discussion of the encoding system used. It is worth noting that the encoding system does not have a direct impact on the performance of systems. In the IBN SINA database, Finglish has been used (See Figure 4). We also used a more standard encoding system: Unicode[2]. For example, the Latin letter "a" which stands for the Arabic letter ālif in Finglish is replaced by the UTF-8 representer of ālif which has the Hex index 0627 in the Unicode table. As previously mentioned, we follow the archigrapheme encoding in the Naskh-3 database (shown in Figure 5). In this encoding, dots are ignored. It is worth noting that dash and brackets referred to in this table will not appear in an actual transliteration. Similarly, these Latin representers are replaced by Unicode representers. For example, Unicode representer 066E is used for the dot-less bā' ٮ (B in archigrapheme encoding) which represents all bā'-like letters in archigrapheme encoding (see Figure 6 to see the correspondence between dot-less bā' and bā'-like letters).

## 3. TRAINING AND BUILDING THE PROPOSED OSR SYSTEM

According to Figure 1, the sub-word labels are first converted to binary descriptors, and then SVMs are used for learning their behavior. The details are provided in the following subsections.

---

[2] http://unicode.org/charts/PDF/U0600.pdf

| Transformation | $T_1$ | $T_2$ | $T_3$ | $T_4$ | $T_5$ | $T_6$ |
|---|---|---|---|---|---|---|
| # elements | 3 | 3 | 2 | 1 | 1 | 4 |
| 1 | BHoleCon | EBPCon | BDotUpFlag | EDotFlag | DRelVertCMDot | ESShapeFlag |
| 2 | BEPCon | EEPCon | BDotDownFlag | | | EClockwise |
| 3 | BBPCon | ERelVertCMEP | | | | EAboveItsBP |
| 4 | | | | | | EBelowItsBP |

Table 1: The various feature vectors associated to BPs, EPs, and SPs of a sub-word.

$$\begin{aligned}(\text{AR,} \quad &\text{HorizFreq,} \quad \text{VertCMRatio,} \quad \cdots, \quad \text{DottedFlag,} \quad \text{BHoleCon}_1, \quad \text{BEPCon}_1, \quad \text{BBPCon}_1, \\ &\text{BHoleCon}_2, \quad \text{BEPCon}_2, \quad \text{BBPCon}_2, \quad \cdots, \quad \text{BHoleCon}_6, \quad \text{BEPCon}_6, \quad \text{BBPCon}_6, \\ &\text{EBPCon}_1, \quad \text{EEPCon}_1, \quad \text{ERelVertCMEP}_1, \quad \cdots, \quad \text{EBPCon}_6, \quad \text{EEPCon}_6, \quad \text{ERelVertCMEP}_6, \\ &\text{BDotUpFlag}_1, \quad \text{BDotDownFlag}_1, \quad \cdots, \quad \text{BDotUpFlag}_6, \quad \text{BDotDownFlag}_6, \\ &\text{EDotFlag}_1, \quad \cdots, \quad \text{EDotFlag}_6, \quad \text{DRelVertCMDot}_1, \quad \cdots, \quad \text{DRelVertCMDot}_6, \\ &\text{ESShapeFlag}_1 \quad \text{EClockwise}_1 \quad \text{EAboveItsBP}_1 \quad \text{EBelowItsBP}_1 \quad \cdots, \\ & \qquad\qquad\qquad\qquad\qquad\qquad\qquad \text{ESShapeFlag}_6 \quad \text{EClockwise}_6 \quad \text{EAboveItsBP}_6 \quad \text{EBelowItsBP}_6)\end{aligned}$$

Table 3: A typical sub-word feature vector composed of feature vectors in Tables 1 and 2. It has 92 elements.

Figure 4: The Finglish encoding table.

Figure 5: The archigrapheme encoding used in this work. A dash before a Latin letter in this table means that that letter could only appear at the end of a sub-word. Brackets around a letter indicates that this letter also has the same form in the middle of a sub-word.

Figure 6: An example of archigrapheme encoding: Archigrapheme dot-less bā' replaces all bā'-like letters shown on the right side.

### 3.1 Conversion of string labels to binary descriptors

The binary-descriptor concept refers to a new way of addressing the highly multi-class nature of sub-word labeling by defining a set of letter binary descriptors to redefine the labeling problem. Figure 7 illustrates this concept. If we assume the alphabet has just three letters, ālif, bā' and tā', the possible combination of these letters (ignoring the order) can be visualized as the parts of the leaves in the figure. Non-overlapping parts indicate single-letter sub-words, while overlapping parts correspond to sub-words composed of associated letters. In the binary-descriptor concept, each leaf is considered as a single binary descriptor. Therefore, the original highly multi-class problem can be replaced by an ensemble of binary descriptor classifiers which are easier to learn thanks to the existence of various state-of-the-art classification methods, such as SVMs, for binary descriptors. In [4], only the binary descriptors that check for the presence of letters in sub-words were considered. In this way, the order is completely ignored. In this work, in order to increase the accuracy in recovering the correct order of letters, we add additional binary descriptors to generate implicit clues to the order of letters in the sub-word. For example, in addition to the binary descriptors of the presence of letters in the sub-words, a similar figure to Figure 7 can be considered but now for the first letter of sub-words. The corresponding binary descriptors will learn the presence of letters as the first letter of the sub-words. The same process can be performed for the second, the third, etc. letters of the sub-words. We use six different types of binary descriptors:

1. $P_{L,w}$: For each letter $w$, $P_{L,w}$ determines if that letter is present in the sub-word or not.

2. $P_{T,w}$: For each letter $w$, $P_{T,w}$ determines if more than one instance of that letter is present in the sub-word or not.

3. $P_{1,w}$: For each letter $w$, $P_{1,w}$ determines if that letter is the first letter of the sub-word or not.

4. $P_{2,w}$: For each letter $w$, $P_{2,w}$ determines if that letter is the second letter of the sub-word or not.

5. $P_{3,w}$: For each letter $w$, $P_{3,w}$ determines if that letter is the third letter of the sub-word or not.

6. $P_{S,s}$: For digit $s$, $P_{S,s}$ determines if that digit is 1 or not in the binary representation of the length of the sub-word. For example, for a sub-word with 3 letters, the binary representation is 11. Therefore, $P_{S,1} = 1$, $P_{S,2} = 1$, $P_{S,3} = 0$, and so on. In this work, only $s = 1, \cdots, 4$ are considered.

For example, for a sub-word "lkm", $P_{L,l} = 1$, $P_{L,k} = 1$, $P_{L,m} = 1$, $P_{1,l} = 1$, $P_{2,k} = 1$, $P_{3,m} = 1$, $P_{S,1} = 1$, $P_{S,2} = 1$, and all the other descriptors are negative. In this example, Latin letters are used for the sake of simplicity.

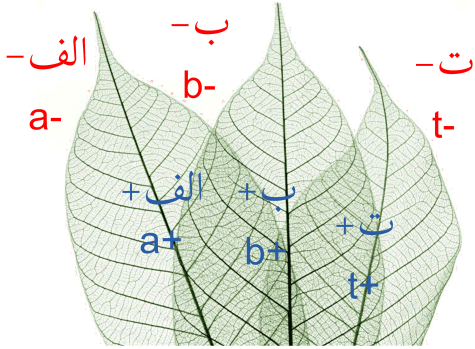

Figure 7: Concept of redefining the sub-words in terms of the letter binary descriptors.

It is worth noting that for some binary descriptors the number of positive samples is very small. However, in order to have a complete system, we trained SVMs on all the binary descriptors. We are working to increase the size of the databases, especially the Naskh database, to include all possible sub-words of any possible length.

### 3.2 Training of SVMs
SVMs classifiers are used to learn the behavior of the binary descriptors. They are a particular type of linear classifier based on the margin-maximization principle [11]. They are powerful classifiers, and have been used successfully in many pattern recognition problems [12]. In [4], we have used SVMs to learn a few binary descriptors with high number of positive samples. Here, we use the same approach to all binary descriptors, trying to strike a balance between positive and negative populations.

A radial basis function (RBF) kernel is used: $k(x_i, x_j) = \exp(-\gamma \|x_i - x_j\|^2)$ where $x_i$ and $x_j$ are two typical feature vectors and $\gamma$ is the kernel parameter. Because all the binary descriptors we have are unbalanced, we use different hyper-parameters: $C$ for controlling the training error impact; $C_+$ for positive samples; and $C_- = C_+/C_j$ for negative samples, where $C_j = n_-/n_+$, with $n_+$ and $n_-$ representing the number of positive and negative samples respectively [13]. The SVM parameters are optimized on the training set to select the best model.

Also, in order to have a probability distribution of the outputs, the following distribution is fitted on the outputs of the trained SVMs:

$$y' = 1/(\sigma_a y + \sigma_b)$$

where $y'$ is distributed between 0 and 1.

### 3.3 Reconstructing the sequence of letters from the binary labels
With the trained SVMs, the system can generate the binary labels of each sub-word. The next step in the recognition process is to reconstruct the candidate sequences out of these labels. First, a set of sequences is built based on the outputs of the $P_{L,w}$ and $P_{T,w}$ descriptors by permuting the positive letters. Then, only those sequences that are compatible with the first letters indicated by the $P_{1,w}$ descriptors are kept. The $P_{S,s}$ are used to select the most probable letters from $P_{L,w}$ and $P_{T,w}$, in order to build the sequences.

The next step is to prune the candidate sequences based on a dictionary. The dictionary for each database was built by extracting all the strings associated with the database sub-words. Therefore, the size of dictionary is equal to the number of unique sub-words (Basis CCs; the BCCs) in the database. It is worth noting that, for the Naskh-3 database, the dictionary is based on archigraphemes. In future, we will use one of the Arabic corpora (for example, the freely available corpus [14]) to build the dictionary.

## 4. EXPERIMENTAL RESULTS
Because of the unbalanced number of positive and negative samples for each descriptor, the classic error rate (ER) is not a suitable measure. Instead, we used the balanced error rate (BER), which is the average of the misclassification rates on examples drawn from positive and negative classes. The BER is defined as follows:

$$\text{ER} = \left( \frac{\text{FN} + \text{FP}}{\text{TP} + \text{FN} + \text{FP} + \text{TN}} \right)$$

$$\text{BER} = \frac{1}{2} \left( \frac{\text{FN}}{\text{TP} + \text{FN}} + \frac{\text{FP}}{\text{FP} + \text{TN}} \right)$$

where FN, TP, FP, and TN represent *false negative*, *true positive*, *false positive*, and *true negative* respectively. In each run, the samples are divided into training and testing subsets. Eighty percent of the samples are considered to be in the training set. Because of the limited number of samples, cross validation has been used and the SVMs are trained to reduce the BER of the test set. The model with the minimum BER is kept as the output of the training process.

Some of the statistics of the two databases are provided in Table 6. The performance of some of the individual binary descriptors for the IBN SINA database is provided in Table 4. The corresponding Latin letters from the Finglish encoding table are also provided in the FNC (Finglish code) column. As can be seen from the table, the complexity of the problem varies for different letters. Also, the number of samples influences the performance of the SVMs. It is worth noting that the number of positive samples is considerably smaller for the first-letter descriptors $P_{1,w}$ compared to $P_L$ descriptors.

| Descriptor Name | FNC | BER | ER | $C_j$ |
|---|---|---|---|---|
| $P_{L,0627}$ | a | 0.18 | 0.20 | 1.64 |
| $P_{L,0628}$ | b | 0.40 | 0.58 | 18.29 |
| $P_{1,0628}$ | b | 0.36 | 0.70 | 30.61 |
| $P_{2,062B}$ | c | 0.06 | 0.12 | 206.06 |

Table 4: The performance statistics of some of the SVMs trained on the IBN SINA dataset.

| Descriptor Name | ARC | BER | ER | $C_j$ |
|---|---|---|---|---|
| $P_{L,0627}$ | A | 0.018 | 0.034 | 14.11 |
| $P_{L,066E}$ | B | 0.063 | 0.10 | 4.14 |
| $P_{1,066E}$ | B | 0.055 | 0.10 | 11.49 |
| $P_{2,0635}$ | C | 0.019 | 0.034 | 14.27 |

Table 5: The performance statistics of the SVMs trained for some of the binary descriptors for the Naskh-3 dataset.

Having the trained SVMs, the OSR system is applied to the sub-words of the database according to Figure 1. The performance of the system is provided in Table 6. The error of letters set (ELS) calculates the average error of the recognized sub-word with respect to the ground truth, ignoring the position of letters in the sequence:

$$ESL = \text{average}_i \left[ 0.5 \{ESL(s_i, \hat{s}_i) + ESL(\hat{s}_i, s_i)\} \right]$$

where $s_i$ and $\hat{s}_i$ are a recognized sub-word and its associated ground truth of the $i^{th}$ sub-word in the manuscript. $ESL(s_1, s_2)$ gives the error of the letters-set of $s_1$ with respect to $s_2$. In contrast, the recognition rate calculated provides the percentage of correctly labeled sub-words in the test set. The recognition rate of the first rank is equal to the all-ranks recognition rate thanks to the presence of first-letter descriptors in the system. The performance of the OSR system on the Naskh-3 database is also provided in Tables 5 and 6. The archigrapheme codes are shown in the ARC (archigrapheme code) column. We are working to achieve high performance by adding sub-words containing more than 3 letters to the database.

## 5. DISCUSSIONS

| Dataset Name | IBN SINA | Naskh-3 |
|---|---|---|
| Number of sub-words (CCs) | 27709 | 2920 |
| Size of sub-words dictionary | 1629 | 2887 |
| Error in letters set (ELS) | 2.54 | 0.18 |
| Recognition rate: first rank[(*)] | 45.74 | 51.83 |
| Recognition rate: first rank | 88.28 | 95.59 |
| Recognition rate: all ranks | 89.66 | 96.26 |

Table 6: Statistics and performance of the proposed sub-word OSR system for the IBN SINA dataset and the Naskh-3 dataset. [(*)]Without considering first-, second- and third-letter descriptors.

We can conclude from Tables 4, 5 and 6 that the binary labels have been learned with a high level of performance (especially in the case of the Naskh-3 database, with as low as 0.034 percent error in letter recognition). Although the performance on the IBN SINA database is good, its lower performance may be associated with degradation of the input images, and also the limited number of samples. Improvement of the skeleton-based features would provide a better description of the sub-words, potentially reducing the possibility of error at the letter-recognition level.

Because the first-, second- and third-letter binary descriptors are used, there is less difference between the first rank and all rank scores. It is worth noting that in the Arabic language proper, a word's initial letter or letters often serve as grammatical markers, while the subsequent letters are usually markers of the word's particular root meaning of the word.

Finally, it should be noted that, although we use the first-letter binary descriptors in our set of descriptors, the complete set of features of each sub-word is used to learn and identify them. Therefore, the system is free of character-segmentation, and completely different from OCR methods.

## 6. CONCLUSIONS AND FUTURE PROSPECTS

A prototype Optical Shape Recognition system is developed that can provide the labels at the sub-word level. The system is able to recognize Arabic sub-words of the scripts on which the system has been trained. In order to avoid line/word segmentation, and also to avoid highly multi-class classification, equivalent binary descriptors are used. SVMs are trained to learn and classify these descriptors, and the outputs of the trained SVMs are combined to recover the original sequence of each sub-word letters. Also, the skeleton-based features used to describe the sub-words are robust with respect to possible variations in the size and direction of the strokes. The system has been separately trained/tested on two databases of different scripts. The second database is a synthesized database based on output from the ACE font layout engine for the Naskh style.

Generalization of the system to generate the manuscript text is under consideration. We are also working on completing the Naskh-style database. Investigation of more descriptive skeleton-based features is yet another goal. We are also considering combining our system with others, such as HMM and two-dimensional measures, in order to benefit from different paradigms and improve the system. Evaluation of the method on other databases, such as IFN/ENIT, is under progress.

**Acknowledgment**
The authors thank NSERC of Canada for their financial support.